\documentclass[times,twocolumn,final]{elsarticle}

\usepackage{prletters}
\usepackage[numbers]{natbib}
\usepackage{float,graphicx,caption,subcaption}
\usepackage{algorithm,algorithmic}
\usepackage{amsmath,amssymb}
\usepackage[colorlinks,linkcolor=red,hyperindex,bookmarks=false]{hyperref}
\biboptions{sort&compress}

\begin{document}

\clearpage

\ifpreprint
  \setcounter{page}{1}
\else
  \setcounter{page}{1}
\fi

\begin{frontmatter}

\title{Real-MFF: A large realistic multi-focus image dataset with ground truth}

\author[1,2]{Juncheng \snm{Zhang}} 
\author[1,2]{Qingmin \snm{Liao}}
\author[3]{Shaojun \snm{Liu}\corref{cor1}} 
\cortext[cor1]{Corresponding author.}
\ead{liusj14@tsinghua.org.cn}
\author[1,2]{Haoyu \snm{Ma}} 
\author[1,2]{Wenming \snm{Yang}} 
\author[4]{Jing-Hao \snm{Xue}}

\address[1]{Department of Electronic Engineering, Tsinghua University, China}
\address[2]{Shenzhen International Graduate School, Tsinghua University, China}
\address[3]{Department of Electronic and Computer Engineering, Hong Kong University of Science and Technology, China}
\address[4]{Department of Statistical Science, University College London, U.K.
}

\received{}
\finalform{}
\accepted{}
\availableonline{}
\communicated{}

\begin{abstract}
Multi-focus image fusion, a technique to generate an all-in-focus image from two or more partially-focused source images, can benefit many computer vision tasks. However, currently there is no large and realistic dataset to perform convincing evaluation and comparison of algorithms in multi-focus image fusion. Moreover, it is difficult to train a deep neural network for multi-focus image fusion without a suitable dataset. In this letter, we introduce a large and realistic multi-focus dataset called Real-MFF, which contains 710 pairs of source images with corresponding ground truth images. The dataset is generated by light field images, and both the source images and the ground truth images are realistic. To serve as both a well-established benchmark for existing multi-focus image fusion algorithms and an appropriate training dataset for future development of deep-learning-based methods, the dataset contains a variety of scenes, including buildings, plants, humans, shopping malls, squares and so on. We also evaluate 10 typical multi-focus algorithms on this dataset for the purpose of illustration. 
\end{abstract}

\begin{keyword}
\KWD Image fusion 
\sep multi-focus images 
\sep multi-focus dataset 
\sep deep learning
\end{keyword}

\end{frontmatter}

\section{Introduction}\label{s:intro}

For most computer vision tasks, such as object detection and identification, it is desirable to use the in-focus images as input rather than blurred ones. However, due to the limited depth-of-field (DOF) of cameras, it is usually difficult to capture the all-in-focus images directly. Therefore, multi-focus image fusion, a technique to fuse two or more partially-focused source images into an all-in-focus image, is very important in the fields of computer vision and image processing, and has drawn considerable attention in recent years.

Existing multi-focus image fusion methods can be roughly categorized into three groups: transform-domain-based methods, spatial-domain-based methods, and deep-learning-based methods. 

Transform-domain-based methods usually first decompose the source images in a transform domain, then fuse the features in the transform domain, and finally reconstruct the all-in-focus image. Laplacian pyramid (LP) \cite{LP}, ratio of low-pass pyramid (RP) \cite{RP}, curvelet transform (CVT) \cite{CVT}, discrete wavelet transform (DWT) \cite{DWT}, dual-tree complex wavelet transform (DTCWT) \cite{RDTCWT}, non-subsampled contourlet transform (NSCT) \cite{NSCT}, principal component analysis (CPA) \cite{Rcpa}, and sparse representation-based methods \cite{SR,nsctsr,SPpr} have been explored to build transform-domain-based methods.

Spatial-domain-based methods can be further classified into three sub-groups: block-based methods, region-based methods and pixel-based methods. The block-based methods \cite{block} first divide images into blocks, then calculate the focus measure of each block, and finally choose the block with the highest focus measure as the corresponding block in the fusion result. Consequently, these algorithms are often affected by the granularity of block partitioning. The region-based methods \cite{region} first segment the input images and then fuse the focused segments of these input images. Therefore, the fusion results of region-based methods highly rely on the segmentation accuracy. The pixel-based methods \cite{DSIFT} calculate the focus measure and fuse the images at the pixel level. Usually, pixel-based methods often produce poor results near the boundary between a focused area and a defocused area.

\begin{figure*}
\centering
\includegraphics[width=0.75\textwidth]{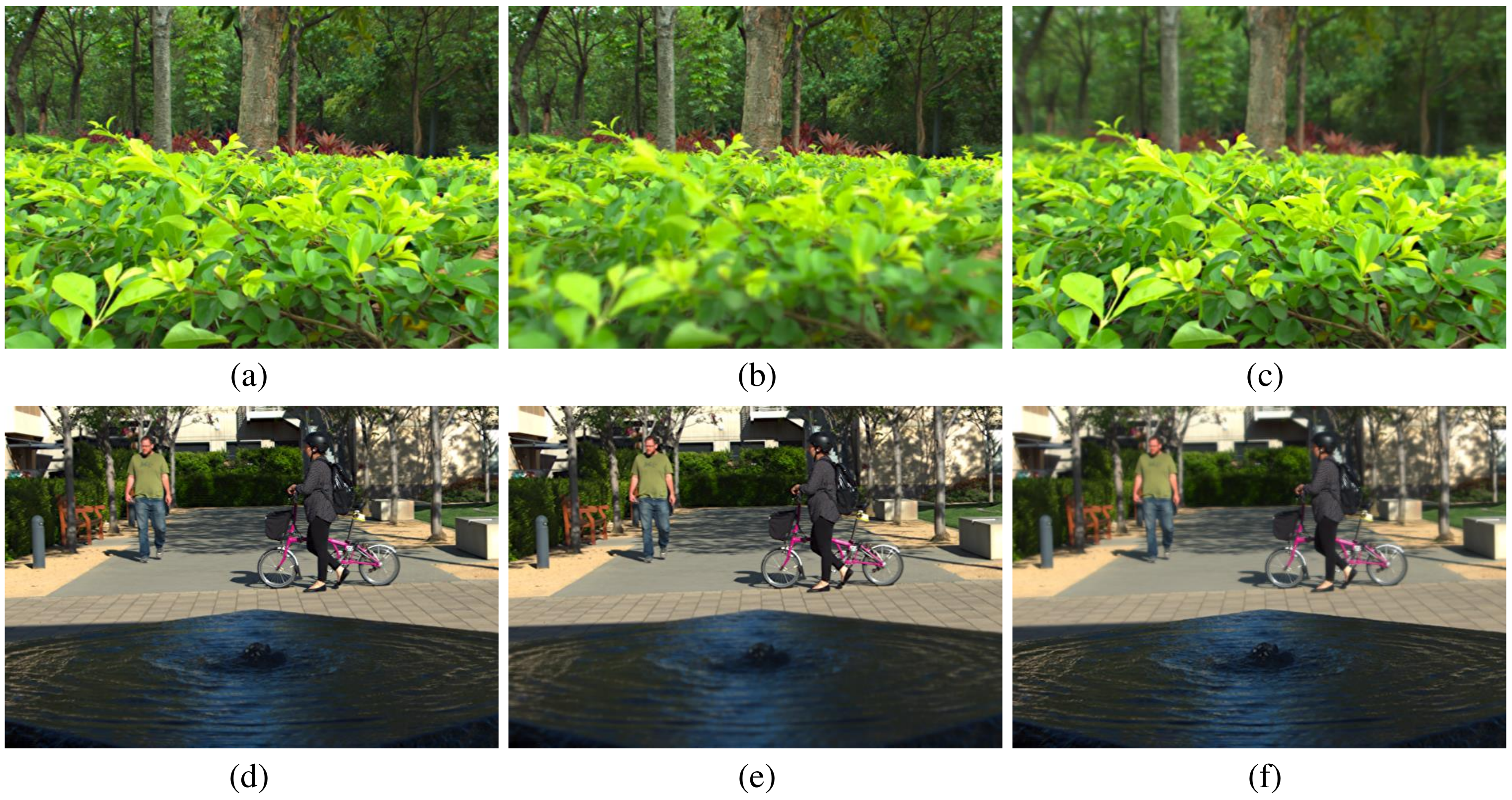}
\caption{Examples in our dataset. (a) and (d) are the source images focused on the foreground; (b) and (e) are the source images focused on the background; and (c) and (f) are the all-in-focus images. The upper row is for a complex example and the lower row is for a simple example.}
\label{fig1}
\end{figure*}

\begin{table*}
\centering
\caption{Characteristics of various datasets.}\label{table1}
\begin{tabular}{lllll}
\hline
Dataset & Data generation method & size & realistic & ground truth\\
\hline
Lytro \cite{lytrodata} & Captured by light field camera & 20 pairs, 520$\times$520 & Yes & No \\
CNN \cite{cnn} & Synthetically generated based on the ImageNet dataset & 1,000,000 pairs, 16$\times$16 & No & Yes \\
BA-Fusion \cite{icme2019ours}  & Synthetically generated based on the Matting dataset &  2,268,000 pairs, 16$\times$16 & No & Yes \\
FuseGan \cite{fusegan} & Synthetically generated based on segmentation datasets & 5,850 pairs, 320$\times$480 & No & Yes \\
Our Real-MFF & Captured by light field camera & 710 pairs, 625$\times$433 & Yes & Yes \\ 
\hline
\end{tabular}
\end{table*}

In the past several years, many deep learning methods have been proposed for multi-focus image fusion. Liu \textit{et al.} \cite{cnn} used a deep convolutional neural network (CNN) to generate a decision map just like the pixel-based methods, and then did some post-processing to produce a final decision map. FuseGan \cite{fusegan} used a generative adversarial network (GAN) to generate decision map. Such methods can be regarded as the network-based implementations of the spatial-domain-based methods. Different from them, Wen \textit{et al.} \cite{ecnn} designed an end-to-end neural network for image fusion.

One of the bottlenecks of using deep neural networks to solve multi-focus fusion problems is the lack of suitable large and realistic database with ground truth for the network training. The most widely used dataset in this area is the Lytro Multi-focus dataset \cite{lytrodata}, which contains 20 pairs of multi-focus images with size 520$\times$520 pixels. This dataset is very small and has no ground truth, therefore, it cannot be used for training a deep neural network. To break through this bottleneck, there are some valuable trials. Liu \textit{et al.} \cite{cnn} used high-quality natural images blurred by Gaussian filters with five different levels of blur to generate a dataset. The images generated by this method were not as real as naturally defocused images, because they were blurred with a spatially invariant defocus kernel and therefore lacked defocus changes. \cite{fusegan,ecnn} used segmentation datasets with manually labeled segmentation as the ground truth map. The ground truth was used as a 0-1 mask map, with ‘0’ for foreground and ‘1’ for background. The foreground and background were then blurred by Gaussian filters separately and finally merged together. In our previous work of BA-Fusion \cite{icme2019ours}, we generated a dataset based on a matting dataset. We chose matted object as the foreground object and high-quality picture as the background. However, the images generated by these methods do not follow the real defocus model and thus need further improvement.

In this letter, we propose a new large and realistic dataset for multi-focus image fusion. The dataset, called Real-MFF, consists of various natural multi-focus images with ground truth, generated by light field images. Fig.~\ref{fig1} shows two  pairs of partially-focused source images and their all-in-focus ground-truth images in our dataset for examples. The contributions of our work can be summarized as follows.

Firstly, we construct a new large and realistic multi-focus dataset, which contains 710 pairs of images that can be used for training deep neural networks. Each pair of images contains two partially-focused images as the source images and an all-in-focus image as the ground truth. The dataset is generated using a light field camera: Lytro illum camera. The source images are produced by choosing different focus planes. 

Secondly, to the best of our knowledge, our dataset is the first large and realistic dataset that can serve as a test bench for validating multi-focus image fusion methods.

Finally, the proposed dataset is both large and realistic so that it can benefit the development of deep-learning-based multi-focus image fusion methods.

\section{Related Work}\label{s:related}

\subsection{Multi-focus image fusion dataset}

As mentioned in Section~\ref{s:intro}, Lytro \cite{lytrodata}, currently the most widely used multi-focus image fusion dataset, only has 20 pairs of images without ground truth. Therefore, it cannot be used for training of deep neural networks. Alternatively, many deep-learning-based algorithms \cite{cnn,ecnn,fusegan,icme2019ours} generated datasets artificially as the training set. But these generated datasets are not natural or realistic, especially near the focused/defocused boundary. The characteristics of these datasets are summarized in Table~\ref{table1}. 

\subsection{Light field image processing}

Different from the conventional cameras, the light field camera records the complete light field information in one shot, and generates images afterwards. In 2005, Ng proposed a Fourier Slice Photography Theorem \cite{Ng}, which proved that images focused at different depths can be computed from a single light field image. Dansereau \textit{et al.} \cite{Donald2015} designed a 4D hyperfan shaped band-pass filter in the frequency domain, which can control the range of focused depths of images in the spatial domain.

\section{Dataset}\label{s:dataset}

\subsection{Image capture and categories}
We use the Lytro illum camera to capture light field images at different places, such as square, campus, shopping mall and street. Photo categories include people, plants, buildings, objects, etc. In addition to the photos taken by us, we also selected 443 images from the Stanford database \cite{datasetsf}, which mainly include some scenes with obviously separable foreground and background, especially those with more complex boundary between foreground and background.

\subsection{Pre-processing}

\begin{figure}[htbp]
\centering
\includegraphics[width=0.49\textwidth]{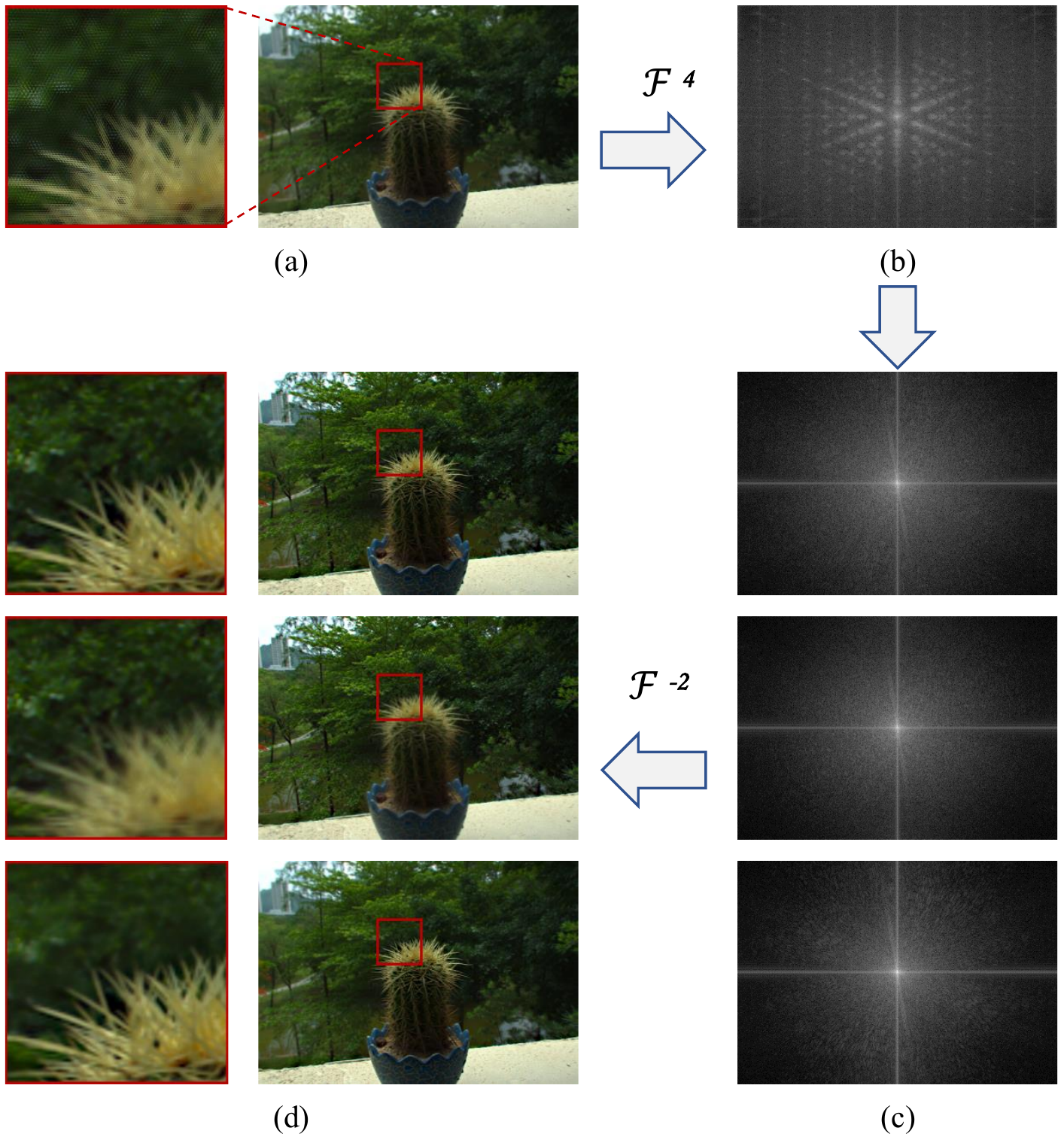}
\caption{Refocusing. (a) is the 4D light field captured by a Lytro camera and (b) is the corresponding 4D Fourier spectrum. (c) shows three 2D slices of the 4D Fourier spectrum, and (d) demonstrates the images generated via applying inverse 2D transforms on (c).
}
\label{FIG2}
\end{figure}

We use the \emph{LFLytroDecodeImage} function in the light field toolbox (LFtoolbox) \cite{Donald2013} to decode light field data and obtain the raw 4D light field data as well as the 2D Lenslet images. A two-dimensional representation of 4D light field data is shown in Fig.~\ref{FIG2}(a). 

\begin{figure*}[htbp]
\centering
\includegraphics[width=0.80\textwidth]{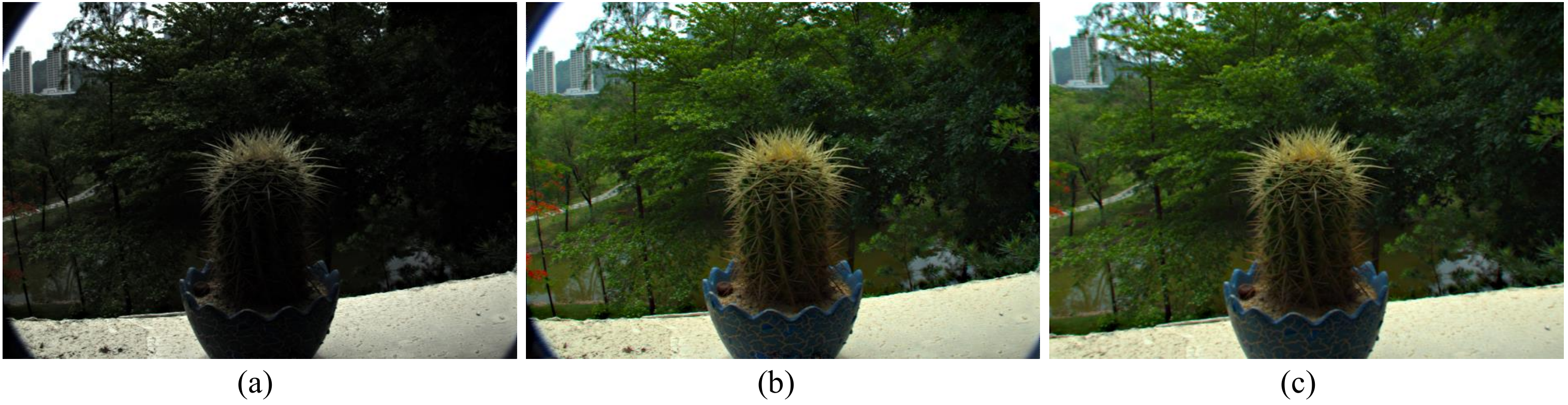}
\caption{Example for pre-processing. (a) is the original image before pre-processing, (b) is the intermediate image after color correction and (c) is the final image after color correction and distortion correction.}
\label{FIG3}
\end{figure*}

Then, we use the \emph{LFdisp} function to generate 2D images. However, the images generated by LFtoolbox are with color deviation and shape distortion effects. We fix the color difference problem through the gamma transform; and with the help of a calibration plate, the problem of shape distortion is solved. 
An illustrative example is shown in Fig.~\ref{FIG3}. In Fig.~\ref{FIG3}(a), the image before color correction is dim and lacks color details; in Fig.~\ref{FIG3}(b), the problem has been fixed after color correction. However, it still can be seen that the object shape in Figs.~\ref{FIG3}(a) and (b) has obvious distortions (e.g.~the table is curved). After distortion correction, the table becomes straight in Fig.~\ref{FIG3}(c). 

\subsection{Refocusing}
The main processing of the dataset is refocusing. For better visualization, we illustrate the processing with an example in Fig.~\ref{FIG2}. Firstly, a 4D light field image (Fig.~\ref{FIG2}(a)) is transformed to a 4D Fourier spectrum space (Fig.~\ref{FIG2}(b)):
\begin{equation}
F(\Omega_u,\Omega_s,\Omega_v,\Omega_t) = DFT( f(u,s,v,t) ),
\end{equation}
where $f$ denotes the 4D light field image, $DFT$ represents the discrete Fourier transform, $F$ is 4D Fourier spectrum, $(u,s,v,t)$ are the two-plane parameterizations of light rays, and $(\Omega_u,\Omega_s,\Omega_v,\Omega_t)$ denotes the corresponding coordinates in the 4D Fourier spectrum space.

According to the Fourier Slice Photography Theorem~\cite{Ng}, we then use 4D hyperfan, \emph{LFBuild4DFreqHyperfan} in LFtoobox~\cite{Donald2013}, as a high-dimensional passband filter to filter the 4D Fourier spectrum (Fig.~\ref{FIG2}(c)):
\begin{equation}
F_{depth} = H_{HF}(\Omega_u,\Omega_s,\Omega_v,\Omega_t,	\theta_1,\theta_2)F(\Omega_u,\Omega_s,\Omega_v,\Omega_t),
\end{equation}
where $H_{HF}$ is the 4D hyperfan passband filter, $F_{depth}$ means the 4D Fourier spectrum of 2D image focused on the specific depth, and parameters $\theta_1$ and $\theta_2$ are the slope parameters of the 4D hyperfan. 

Finally, the 2D images (Fig.~\ref{FIG2}(d)) can be generated by the inverse Fourier transform of the processed 4D Fourier spectrum: 
\begin{equation}
I = IDFT(F_{depth}),
\end{equation}
where $IDFT$ represents inverse discrete Fourier transform, and $I$ denotes the 2D image focused on the specific depth.

Since the depth range usually changes for different scenes, we need to first manually select the specific depth range of focus in the Lytro software, then record the depth of focus for each image, and finally generate the desired image focused on different depth ranges. Therefore, $\theta_1$ and $\theta_2$ usually change for different images. Typically, both $\theta_1$ and $\theta_2$ are in the range of [-2, 2].
The pseudocode of the algorithm is shown in Algorithm~\ref{alg1}.

\begin{algorithm}
\caption{Refocusing} 
\label{alg1}
\begin{algorithmic}
\FOR{each light field image} 
\STATE Decode the light field data: $f$;
\STATE Transform $f$ into the frequency domain by FFT:
\STATE $F = DFT(f)$;
\STATE $j$ = the depth at foreground;
\STATE $k$ = the depth at background;
\STATE $l = (j+k)/2$ ;
\STATE \# Generate the foreground focused image
\STATE Build the 4D hyperfan with slopes $j$ and $l$:
\STATE $H_{FG} = H_{HF}(j,l)$;
\STATE $F_{FG} = H_{FG}F$;
\STATE $I_{FG} = IDFT(F_{FG})$;
\STATE \# Generate the background focused image
\STATE Build the 4D hyperfan with slopes $k$ and $l$:
\STATE $H_{BG} = H_{HF}(k,l)$;
\STATE $F_{BG} = H_{BG}F$;
\STATE $I_{BG} = IDFT(F_{BG})$;
\STATE \# Generate the all-in-focus image
\STATE Build the 4D hyperfan with slopes $j$ and $k$:
\STATE $H_{AF} = H_{HF}(j,k)$;
\STATE $F_{AF} = H_{AF}F$;
\STATE $I_{AF} = IDFT(F_{AF})$;
\ENDFOR
\end{algorithmic}
\end{algorithm}

\section{Experiments}

\subsection{Dataset}

\begin{figure}[h]
\centering
\includegraphics[width=0.48\textwidth]{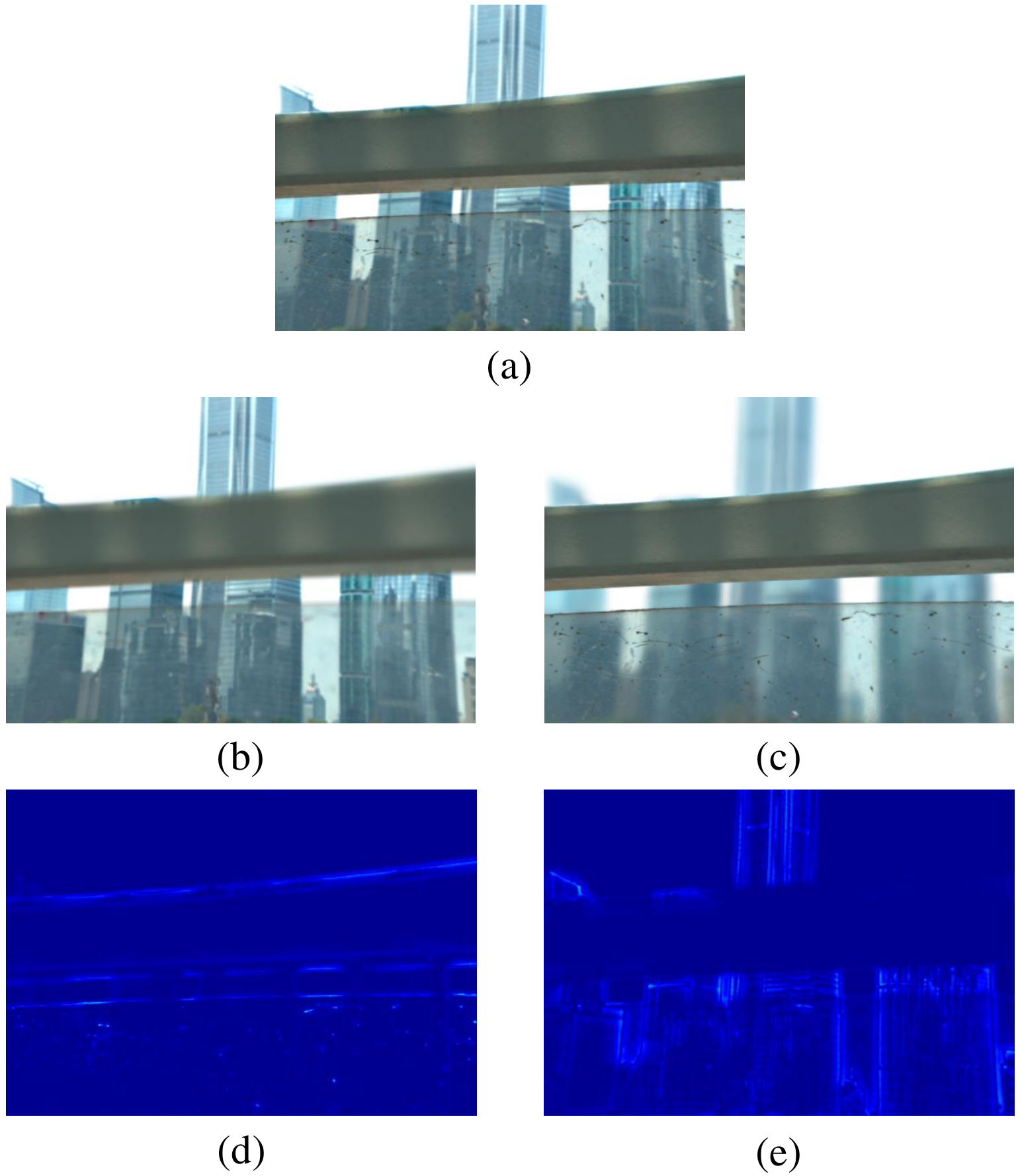}
\caption{Example of the generated images. (a) is the all-in-focus image; (b) is the image focused on the background; (c) is the image focused on the foreground; (d) is the difference map between (a) and (b); and (e) is the difference map between (a) and (c).}
\label{FIG4}
\end{figure}

\begin{figure*}
\centering
\includegraphics[width=0.8\textwidth]{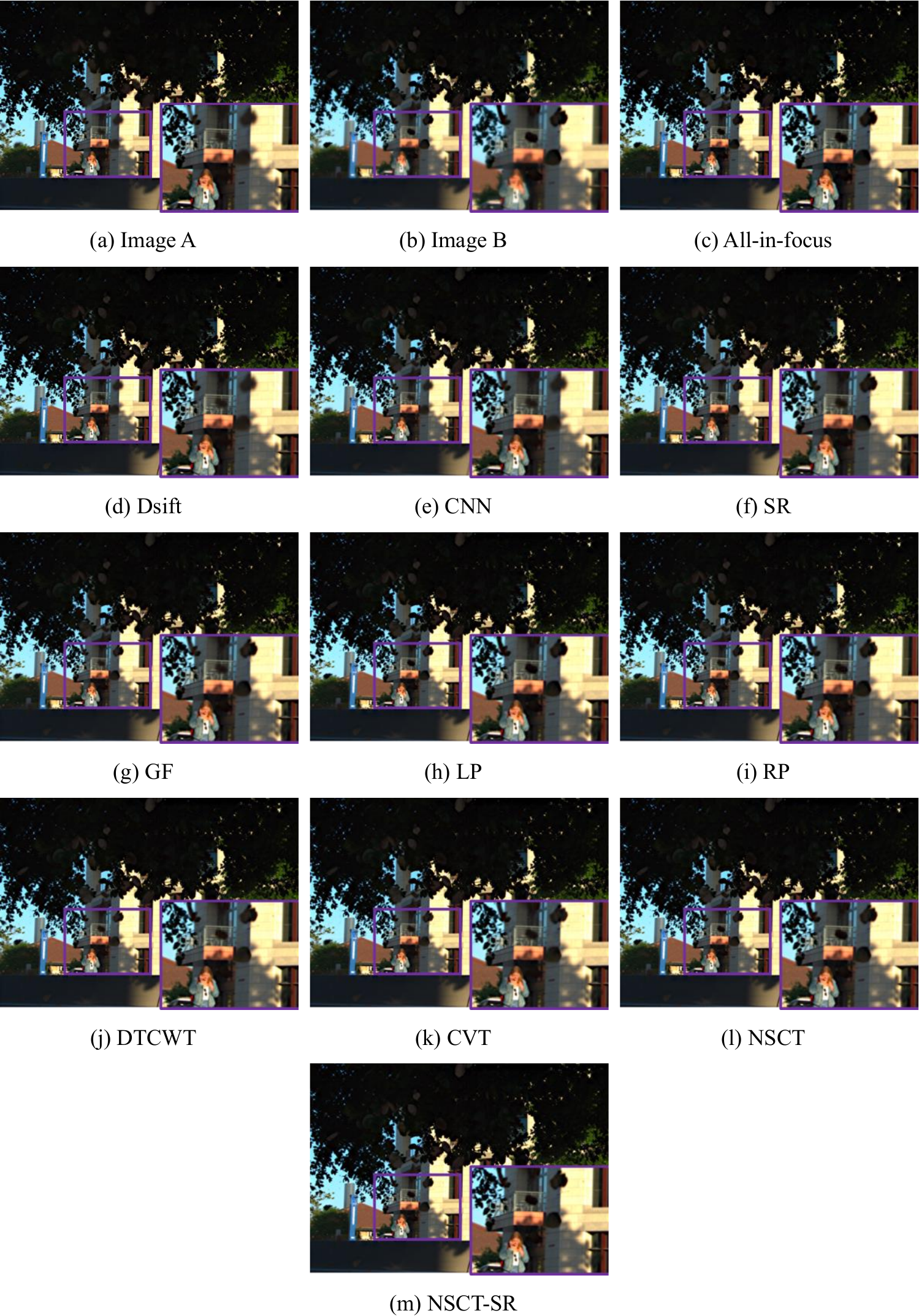}
\caption{Visual comparison. (a) and (b) are the two partially-focused source images; (c) is the ground truth all-in-focus image; and (d)-(m) are the fusion results of the 10 methods.}
\label{FIG5}
\end{figure*}

\begin{figure*}
  \centering
  \begin{subfigure}{0.245\textwidth}
    \includegraphics[width=\textwidth]{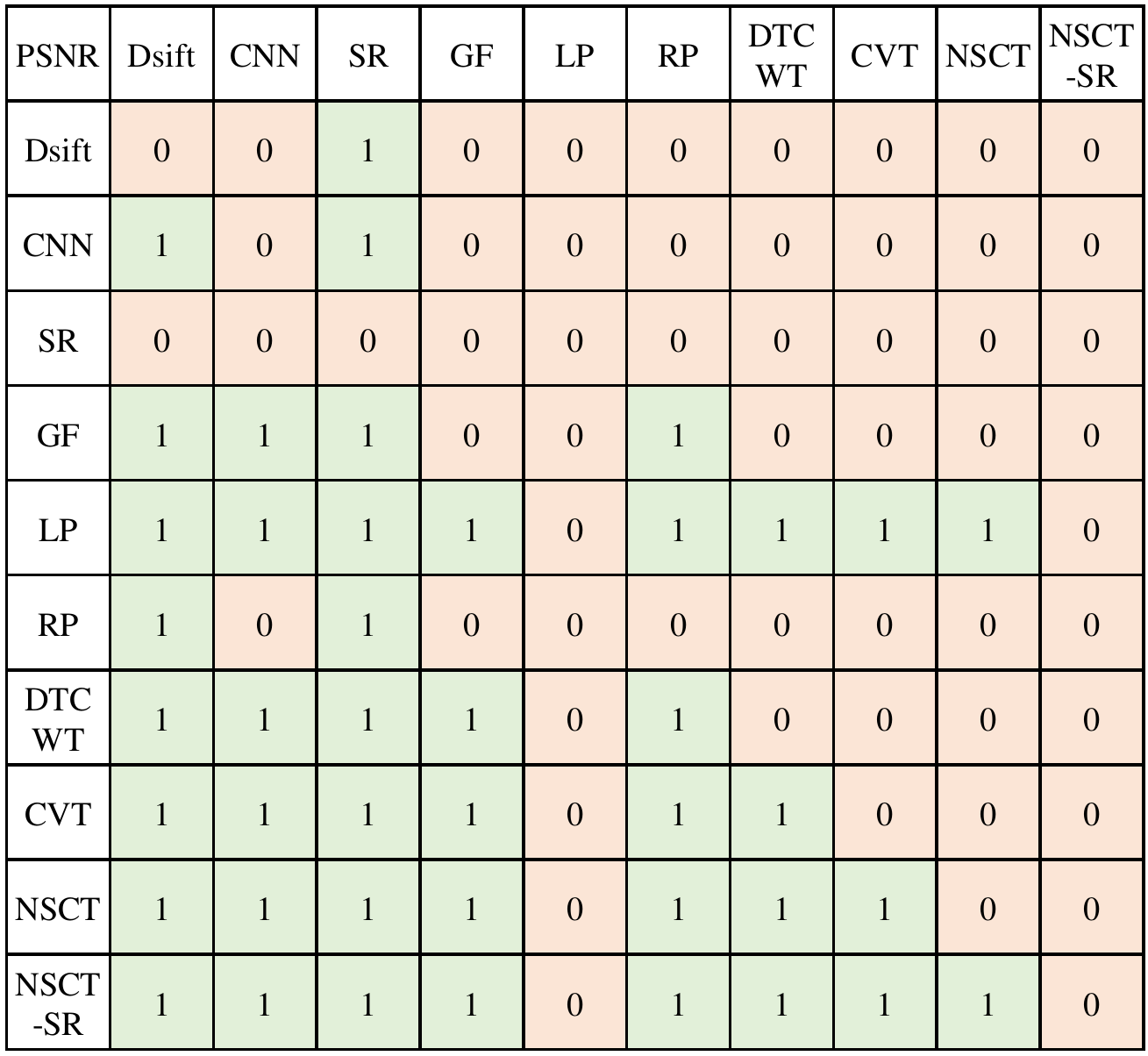}
    \caption{PSNR}\label{fig:matrix:PSNR}
  \end{subfigure}
  \begin{subfigure}{0.245\textwidth}
    \includegraphics[width=\textwidth]{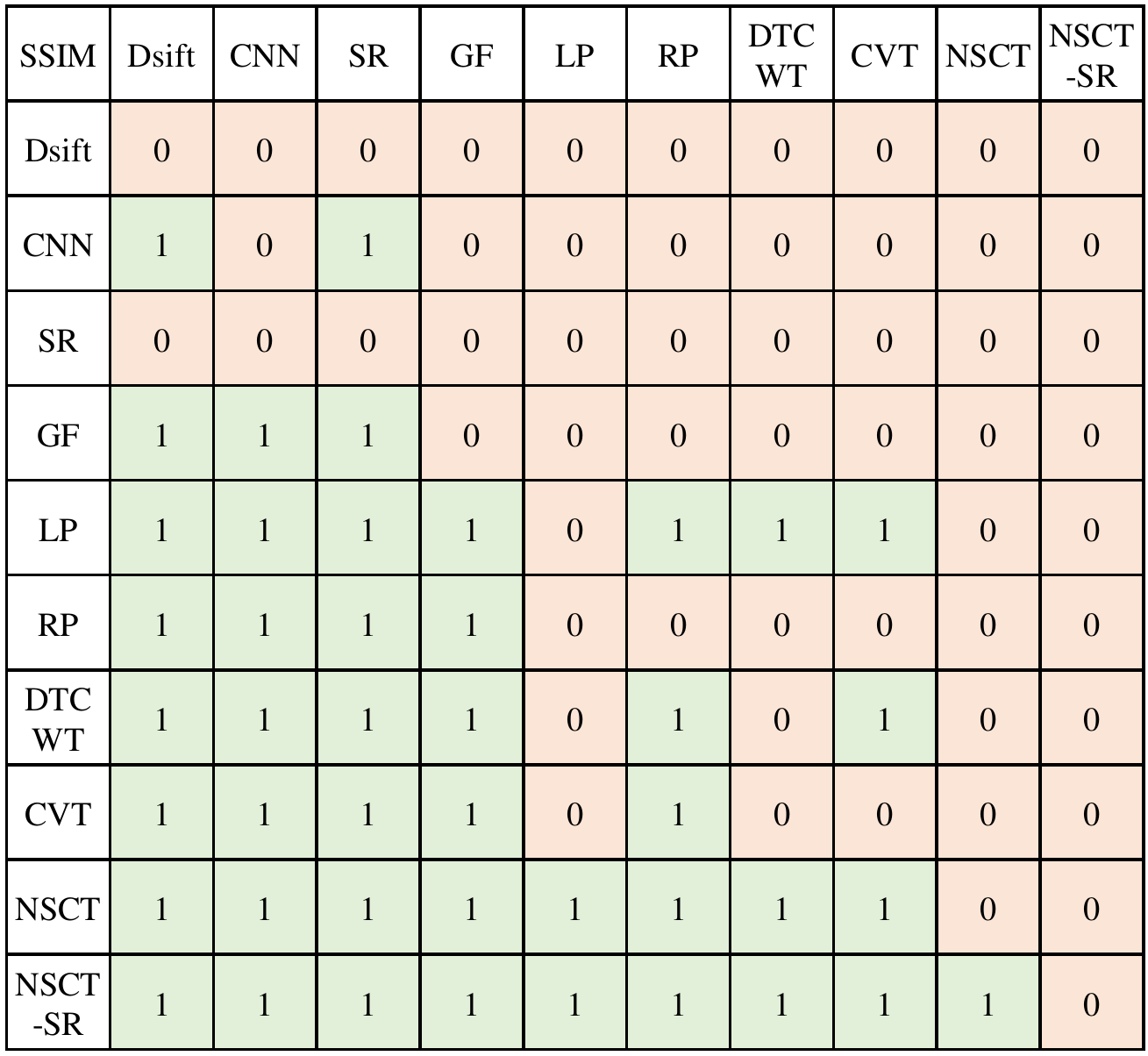}
    \caption{SSIM}\label{fig:matrix:SSIM}
  \end{subfigure}
  \begin{subfigure}{0.245\textwidth}
    \includegraphics[width=\textwidth]{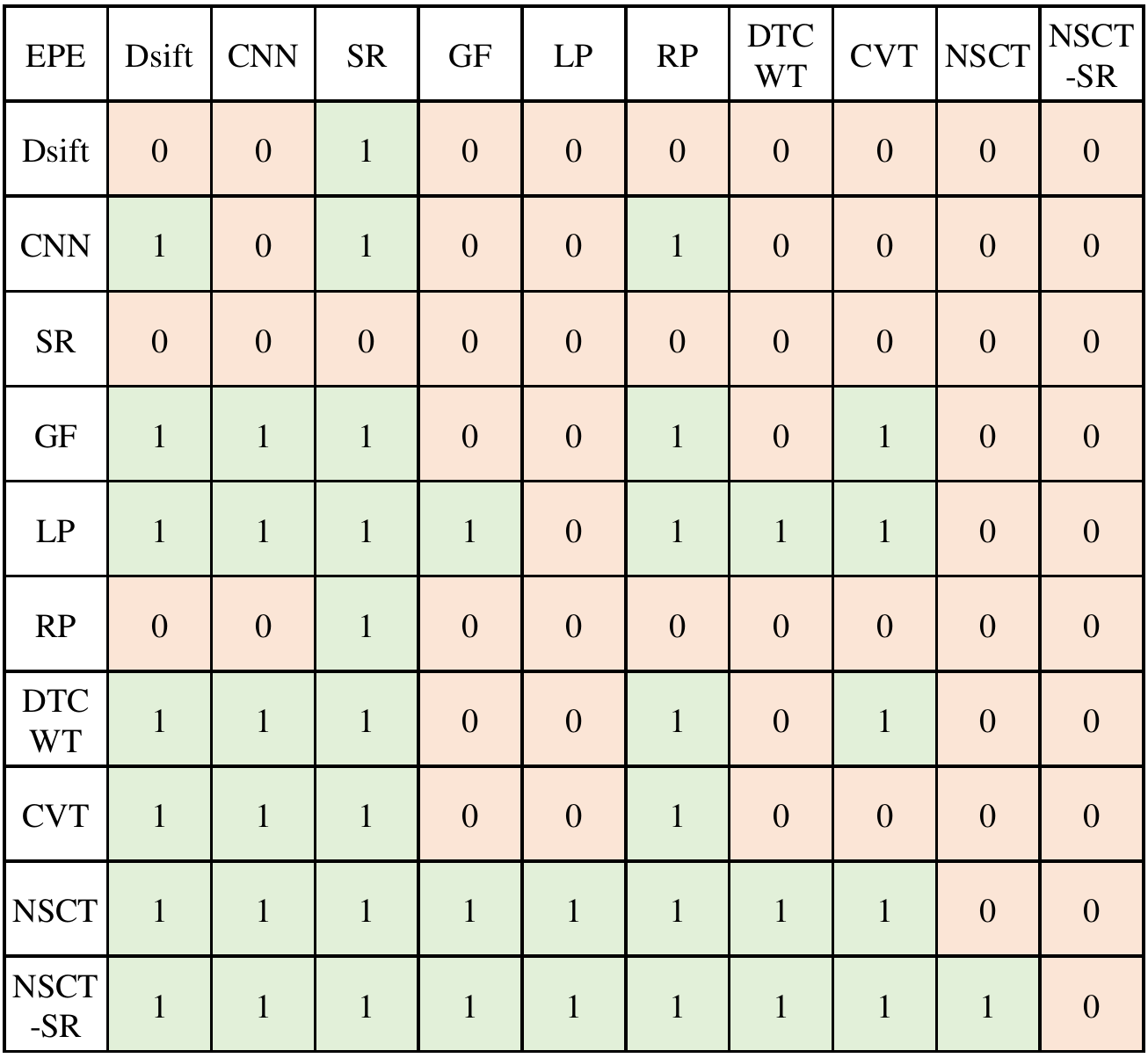}
    \caption{EPE}\label{fig:matrix:EPE}
  \end{subfigure}
  \begin{subfigure}{0.245\textwidth}
    \includegraphics[width=\textwidth]{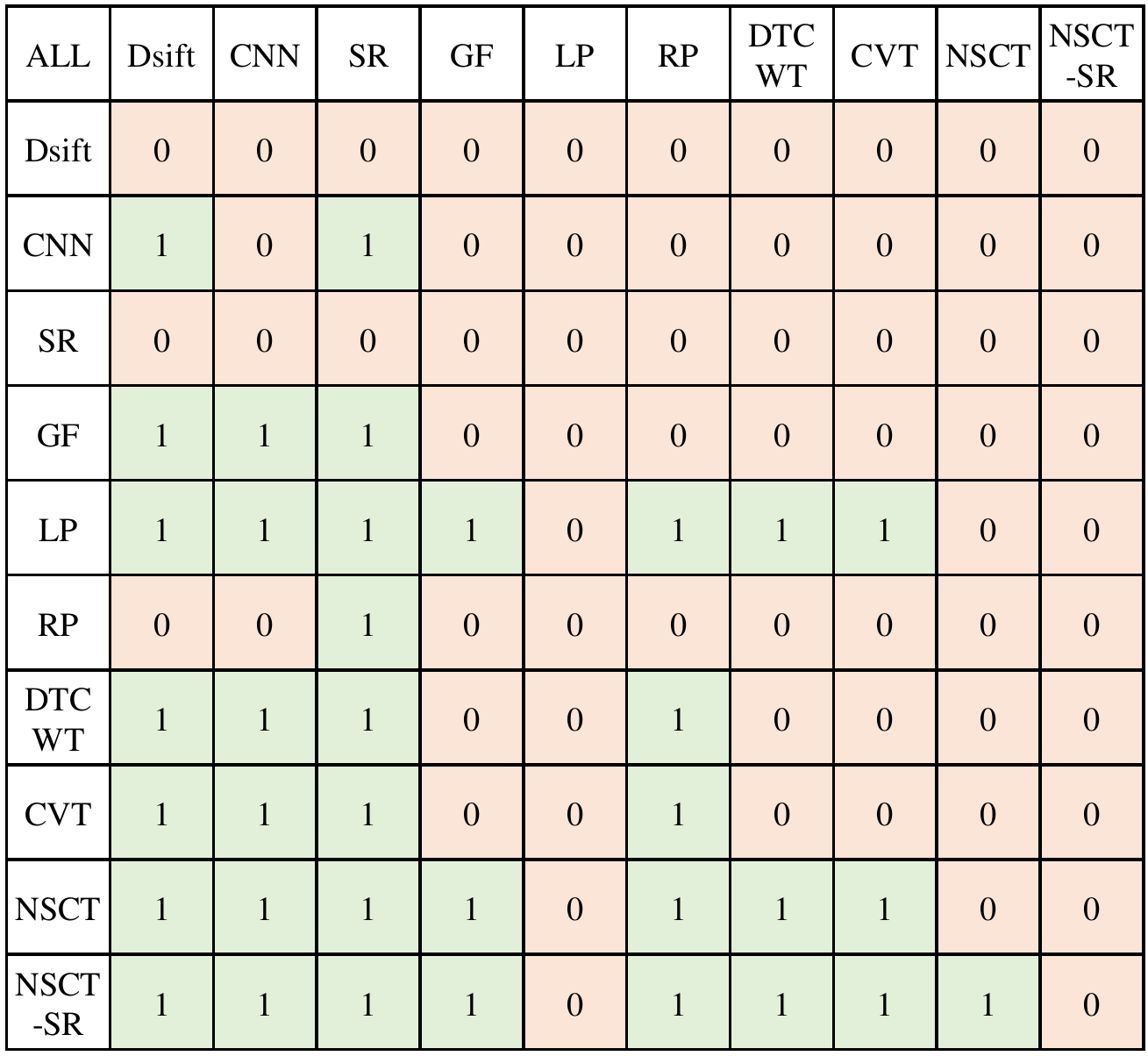}\hfill
    \caption{ALL}\label{fig:matrix:ALL}
  \end{subfigure}
\caption{Significance testing results for the performance of different methods in terms of (a) PSNR, (b) SSIM, (c) EPE and (d) the intersection (denoted by `ALL') of (a)-(c), where `1' indicates that the algorithm of the row is significantly better than the algorithm of the column, and `0' otherwise, at the significance level of 0.05.}
\label{FIG6}
\end{figure*}

Our Real-MFF dataset contains 710 pairs of images, where each pair includes two partially-focused source images focused on the foreground or background respectively, and an all-in-focus image as the ground truth. We hope that the ground truth is similar to the composition of two source images: the foreground region should be similar to the source image focused on the foreground, and the background region should be similar to the source image focused on the background. Fig.~\ref{FIG4} shows an example for the difference maps between the all-in-focus image and the two source images. The difference shown in Fig.~\ref{FIG4}(d) indicates that the ground truth (Fig.~\ref{FIG4}(a)) matches the background of the image focused on the background (Fig.~\ref{FIG4}(b)); and the difference shown in Fig.~\ref{FIG4}(e) indicates that the ground truth (Fig.~\ref{FIG4}(a)) matches the foreground of the image focused on the foreground (Fig.~\ref{FIG4}(c)). 

\begin{table*}[h]
\centering
\caption{Quantitative comparison of multi-focus image fusion methods on our dataset.}\label{table2}
\begin{tabular}{lllllllllll}
\hline
 & Dsift \cite{DSIFT} & CNN \cite{cnn} & SR \cite{SR} & GF \cite{GF} & LP \cite{LP} & RP \cite{RP} & DTCWT \cite{RDTCWT} & CVT \cite{CVT} & NSCT \cite{NSCT} & NSCT-SR \cite{nsctsr}\\
\hline
\hline
\multicolumn{11}{c}{on the whole dataset}\\
\hline
PSNR & 39.96 & 40.46 & 39.68 & 40.74 & 41.32 & 40.37 & 40.86 & 40.90 & 41.18 & 41.32\\
SSIM & 0.9865 & 0.9880 & 0.9857 & 0.9890 & 0.9904 & 0.9891 & 0.9896 & 0.9892 & 0.9905 & 0.9906\\
EPE & 1.959 & 1.901 & 2.060 & 1.865 & 1.790 & 1.939 & 1.876 & 1.886 & 1.762 & 1.736\\
\hline
\hline
\multicolumn{11}{c}{on the simple dataset}\\
\hline
PSNR & 42.71 & 43.41 & 42.13 & 43.45 & 43.93 & 42.88 & 43.46 & 43.52 & 43.72 & 43.87\\
SSIM & 0.9902 & 0.9920 & 0.9888 & 0.9921 & 0.9924 & 0.9914 & 0.9922 & 0.9921 & 0.9926 & 0.9927\\
EPE & 1.446 & 1.375 & 1.526 & 1.357 & 1.294 & 1.408 & 1.349 & 1.349 & 1.268 & 1.244\\
\hline
\hline
\multicolumn{11}{c}{on the complex dataset}\\
\hline
PSNR & 38.39 & 38.76 & 38.26 & 39.18 & 39.81 & 38.92 & 39.35 & 39.38 & 39.72 & 39.85\\
SSIM & 0.9843 & 0.9857 & 0.9838 & 0.9873 & 0.9892 & 0.9878 & 0.9881 & 0.9876 & 0.9894 & 0.9894\\
EPE & 2.255 & 2.205 & 2.368 & 2.159 & 2.076 & 2.245 & 2.181 & 2.196 & 2.048 & 2.020\\
\hline
\end{tabular}
\end{table*}

Furthermore, in order to evaluate the effect of difficulty level on an algorithm, especially for simple and complex boundaries, we manually divide the dataset into a simple set (260 pairs) and a complex set (450 pairs) according to the complexity of focused/defocused boundary. The division is conducted via a three-person majority voting to reduce the bias caused by subjective selection. Two examples are shown in Fig.~\ref{fig1}, where Fig.~\ref{fig1}(a) is a complex case and Fig.~\ref{fig1}(d) is a simple case. For the complex case in Fig.~\ref{fig1}(a), we can find it difficult to draw a curve as the boundary of the foreground and the background, whereas for the simple case in Fig.~\ref{fig1}(d), we can easily take the border of the cistern as the boundary. 

\subsection{Test Bench}

We evaluate 10 typical multi-focus image fusion methods, including some transform-domain-based methods (LP \cite{LP}, RP \cite{RP}, DWT \cite{DWT}, CVT \cite{CVT}, DTCWT \cite{RDTCWT}, NSCT \cite{NSCT}, SR \cite{SR}, NSCT-SR \cite{nsctsr}), some spatial-domain-based methods (Dsift \cite{DSIFT}, GF \cite{GF}), and a deep-learning-based methods (CNN \cite{cnn}), on our dataset. 

We employ the peak signal to noise ratio (PSNR), structure similarity (SSIM) and end point error (EPE) as the evaluation measures, and in terms of these measures, the performances of the 10 multi-focus image fusion methods on the whole dataset, the simple dataset and the complex dataset are listed in Table~\ref{table2}. 
Unsurprisingly, the results of these algorithms on the simple dataset are generally better than those on the complex dataset. The results also show that, among these algorithms, NSCT and NSCT-SR are the top two on our dataset. 

For illustration, we choose an example with complex foreground borders from our dataset, and show the visual results of various methods in Fig.~\ref{FIG5}. We can see the difference among these 10 methods in the regions of the falling leaves and the people in the background. It is hard for the spatial-domain-based methods to generate a precise decision map, which means that the resultant images from these methods have some quite blurry areas. Consequently, their PSNRs are usually lower than those of other methods. For the transform-domain-based methods, complex boundaries will decrease the performance of the algorithms, but fortunately this bad effect will not result in complete failure of the algorithms. Moreover, we can see that CNN and Dsift have blurred results for the falling leaves, while NSCT-SR has better results in this area.

In order to further analyze whether the performances of these methods are statistically significantly different from each other, we perform the Wilcoxon signed rank tests \cite{wilcoxon1945individual} on the experimental results in Table~\ref{table2}. The significance testing results in PSNR, SSIM and EPE are shown in Fig.~\ref{FIG6}, as well as the intersection of these three results (denoted by `ALL'), in all of which the number `1' indicates that the algorithm of the row is significantly better than the algorithm of the column at the significance level of 0.05, and the number `0' otherwise. Based on the results, NSCT-SR performs the best on our dataset.

\section{Conclusions}\label{s:conclusion}
In this letter, we propose a large and realistic dataset for multi-focus image fusion tasks. To the best of our knowledge, our Real-MFF dataset is the first large and realistic dataset for this purpose. The dataset is generated by light field images captured with a light field camera. We use the 4D hyperfan in the frequency domain as a band-pass filter to produce images focused on different depths manually. The proposed dataset can serve as a test bench for evaluating multi-focus image fusion methods; furthermore, it can benefit the development of deep-learning-based methods for multi-focus image fusion in the future. It is also our future work to continuously increase the size, diversity and quality of this dataset.

\section*{Acknowledgement}
We thank the reviewers and editors for their constructive comments that improved our manuscript. This work was supported in part by the National Natural Science Foundation of China (No.61771276). 

\section*{Statement}
This paper has been pubilished on Pattern Recognition Letters.

\bibliographystyle{elsarticle-num}
\bibliography{juncheng}

\end{document}